\documentclass[letterpaper]{article} 
\usepackage{aaai23}  
\usepackage{times}  
\usepackage{helvet}  
\usepackage{courier}  
\usepackage[hyphens]{url}  
\usepackage{graphicx} 
\usepackage{natbib}  
\usepackage{caption} 
\usepackage{amsfonts}       
\usepackage{nicefrac}       
\usepackage{microtype}      %
\usepackage{fancyhdr}       
\usepackage{graphicx}       
\usepackage{amsthm}
\usepackage{amsmath}
\usepackage{booktabs, multirow} 
\usepackage{soul}
\usepackage[table]{xcolor} 
\usepackage{changepage,threeparttable} 

\usepackage{mathtools} 
\usepackage{tikz} 
\usepackage{graphicx,multicol}
\usepackage{caption}
\usepackage{subcaption}
\usepackage{multirow}
\usepackage{enumitem}
\usepackage{comment}
\usepackage{float}

\usepackage[ruled,linesnumbered]{algorithm2e}
\SetKwComment{Comment}{/* }{ */}

\theoremstyle{definition}
\newtheorem{definition}{Definition}[section]
\newcommand{\para}[1]{{\noindent \textbf{#1}}}

\newcommand{\ourapp}{DPAUC}
\newcommand{\ourapprr}{$\text{DPAUC}_{\text{RR}}$}
\newcommand{\ourapplap}{$\text{DPAUC}_{\text{Lap}}$}

\title{DPAUC: Differentially Private AUC Computation in Federated Learning}
\author{
    Jiankai Sun\textsuperscript{\rm 1},
    Xin Yang\textsuperscript{\rm 1},
    Kevin Yao\textsuperscript{\rm 1},
    Junyuan Xie\textsuperscript{\rm 2},
    Di Wu\textsuperscript{\rm 2},
    Chong Wang \textsuperscript{\rm 3}\thanks{Work was done when the author was working at ByteDance Inc.}\\
}
\affiliations{
    \textsuperscript{\rm 1}ByteDance Inc.,
     \textsuperscript{\rm 2}ByteDance Ltd,
      \textsuperscript{\rm 3}Apple \\

    \{jiankai.sun, yangxin.yx, kevin.yao, junyuan.xie, di.wu \}@bytedance.com, mr.chongwang@apple.com
}

\usepackage{bibentry}

\begin{document}

\maketitle

\begin{abstract}
Federated learning (FL) has gained significant attention recently as a privacy-enhancing tool to jointly train a machine learning model by multiple participants. 
The prior work on FL has mostly studied how to protect label privacy during model training. However, model evaluation in FL might also lead to potential leakage of private label information.
In this work, we propose an evaluation algorithm that can  accurately compute the widely used AUC (area under the curve) metric when using the label differential privacy (DP) in FL. Through extensive experiments, we show our algorithms can compute accurate AUCs compared to the ground truth. The code is available at {\url{https://github.com/bytedance/fedlearner/tree/master/example/privacy/DPAUC}}

\end{abstract}

\section{Introduction}
\label{sec:intro}

With increasing concerns over data privacy in machine learning, regulations like CCPA\footnote{California Consumer Privacy Act}, HIPAA\footnote{Health Insurance Portability and Accountability Act}, and GDPR\footnote{General Data Protection Regulation, European Union} have been introduced to regulate how data can be transmitted and used. 
To address privacy concerns, \textit{federated learning}~\cite{mmr+17,hhh+20,ym20,gcyr20} has become an increasingly popular tool to enhance privacy by allowing training models without directly sharing their data. Depending on how data is split across parties, FL can be mainly classified into two categories~\cite{ylct19}: \textit{Horizontal Federated Learning}~\cite{Jonas2020,Jenny2020,Sai2020,Li2020} and \textit{Vertical Federated Learning}~\cite{vgsr18,gr18,akk+20,csm+20}. In Horizontal FL (hFL), data is split by entity (\textit{e.g.} a person), and data entities owned by each party are disjointed from other parties. In Vertical FL (vFL), a data entity is split into different attributes (\textit{e.g.} features and labels of the same person), and each party might own the same data entities but their different attributes. In this paper, we focus on the setting of hFL which enables devices (\textit{i.e.} mobile phones) to collaboratively learn a machine learning model (\textit{i.e.} binary classifier) while keeping all the training and testing data on the device.

Although raw data is not shared in federated learning, sensitive information may still be leaked when gradients and/or model parameters are communicated between parties. In hFL, ~\cite{zhu2019dlg} showed that an honest-but-curious server can uncover the raw features and labels of a device by knowing the model architecture, parameters, and  communicated gradient of the loss on the device's data. Based on their techniques, \cite{zhao2020idlg} showed that the ground truth label of an example can be extracted by exploiting 
the directions of the gradients of the weights connected to the logits of different classes.  Researchers have shown that vFL can still leak data information indirectly. For example, ~\cite{li2021label} demonstrated that the gradient norms and directions 
can leak label information in the two-party vFL setting. However, the prior work on FL privacy mostly focuses on model training and there can also be privacy leaks from model evaluation. Specifically, the private label information owned by clients/devices can be leaked to the server when computing evaluation metrics in FL. 

Previous work ~\cite{aucleak2013} has shown that releasing the actual ROC curves on a private test dataset can allow an attacker with some prior knowledge of the test dataset to recover some sensitive information about the dataset.  
Some recent works\cite{DPRegressionDiag2016,Dp4ML2014}  proposed to provide differential privacy (DP) for plotting and releasing ROC curves. However, they have several challenges such as how to privately compute the true positive rate (TPR) and false positive rate (FPR) values and how many and what thresholds to pick\cite{Dp4ML2014}. And they are not designed and applicable for evaluating FL models. 
As the ground-truth labels often contain highly sensitive information (\textit{e.g.}, whether a user has purchased (in online advertising) or whether a user has a disease or not (in disease prediction)~\cite{vepakomma2018split, li2021label}, it cannot be directly shared between clients and servers, and clients and clients. Hence preventing the label leakage from the AUC computation in the general setting of FL is challenging.

To address the challenge, we consider the area under the  Receiver operating characteristic (ROC) curve as the target AUC to evaluate the accuracy of a binary classifier. Since the class label is often the most sensitive information in a prediction task, our goal in this paper is to achieve \textit{label differential privacy} ~\cite{labeldp21} while plotting the ROC curve.
To this end, we propose to adopt the Laplace mechanism \footnote{Other mechanisms such as the Gaussian mechanism are applicable too)} to add noise to the shared intermediate information between the server and clients to plot the Receiver operating characteristic (ROC) curve  and calculate the area under the ROC curve as  the AUC. We conduct extensive experiments to demonstrate the effectiveness of our proposed approach.

\section{Preliminaries}
\label{sec:preliminaries}

We focus on the setting of hFL which contains one server and multi-clients (devices). The labels are distributed in multi clients and our proposed approach can compute the evaluation metric AUC with label differential privacy. We start by introducing some background knowledge of our work.

\subsection{Label Differential Privacy}

Differential privacy (DP) ~\cite{dmns06,dr14} is a quantifiable and rigorous privacy framework. We adopt the following definition of DP.
We define Differential privacy (DP) ~\cite{dmns06,dr14} as the following:
\begin{definition}[Differential Privacy]
Let $\epsilon,\delta\in \mathbb{R}_{\geq 0}$, a randomized mechanism $\mathcal{M}$ is $(\epsilon,\delta)$-differentially private (i.e. $(\epsilon, \delta)$-DP), if for any of two neighboring training datasets $D,D'$,
and for any subset $S$ of the possible output of $\mathcal{M}$, we have

\begin{align*}
    \Pr[\mathcal{M}(D)\in S]\leq e^{\epsilon}\cdot \Pr[\mathcal{M}(D')\in S]+\delta.
\end{align*}
\end{definition}

If $\delta = 0$,  then $M$ is $\epsilon$-differentially private (i.e. $\epsilon$-DP).

In our work, we focus on protecting the privacy of label information. Following ~\cite{labeldp21}, we define label differential privacy as the following:

\begin{definition}[Label Differential Privacy]
Let $\epsilon,\delta\in \mathbb{R}_{\geq 0}$, a randomized mechanism $\mathcal{M}$ is $(\epsilon,\delta)$-label differentially private (i.e. $(\epsilon,\delta)$-LabelDP), if for any of two neighboring training datasets $D,D'$ \textit{that differ in the label of a single example},
and for any subset $S$ of the possible output of $\mathcal{M}$, we have

\begin{align*}
    \Pr[\mathcal{M}(D)\in S]\leq e^{\epsilon}\cdot \Pr[\mathcal{M}(D')\in S]+\delta.
\end{align*}
\end{definition}

If $\delta = 0$,  then $M$ is $\epsilon$-label differentially private (i.e. $\epsilon$-LabelDP).

Our proposed approach also shares the same setting with local DP ~\cite{localdpDuchi2013,localdprappor2014,localdp2008,localdp2019} which assumes that the data collector (server in our paper) is untrusted. Following the same setting with local DP, in our proposed approach, each client locally perturbs their sensitive information with a DP mechanism and transfers the perturbed version to the  server. After receiving all clients' perturbed data, the server calculates the statistics and publishes the  result of AUC. We define local DP as the following:

\begin{definition}[Local Differential Privacy]
Let $\epsilon > 0$ and $ 1 > \delta \geq 0$, a randomized mechanism $\mathcal{M}$ is $(\epsilon,\delta)$-local differentially private (i.e. $(\epsilon, \delta)$-LocalDP), if and only if for any pair of input values $v$ and $v'$ in domain $D$, and for any subset $S$ of possible output of $\mathcal{M}$, we have

\begin{align*}
    \Pr[\mathcal{M}(v)\in S]\leq e^{\epsilon}\cdot \Pr[\mathcal{M}(v')\in S]+\delta.
\end{align*}
\end{definition}

If $\delta = 0$,  then $M$ is $\epsilon$-local differentially private (i.e. $\epsilon$-LocalDP).

\begin{definition}[Sensitivity]
Let $d$ be a positive integer, $\mathcal{D}$ be a collection of datasets, and $f: \mathcal{D} \rightarrow \mathcal{R}^d$ be a function. The sensitivity  of a function, denoted $\Delta f$, is defined by $\Delta f = max ||f(D)-f(D')||_p
$ where the maximum is over all pairs of datasets $D$ and $D'$ in $\mathcal{D}$ differing in at most one element and  $||\cdot||_p
$ denotes the $l_p$ norm.

\end{definition}




\begin{definition}[Laplace Mechanism ]  Laplace mechanism defined by ~\cite{dpdwork2014} preserves $(\epsilon, 0)$-differential privacy if the random noise is drawn from the Laplace distribution with parameter $\Delta/\epsilon$, where $\Delta$ is the $l_1$ sensitivity and $\epsilon$, is the corresponding privacy budget.
\end{definition}

In this paper, we leverage two DP properties ~\cite{dpdwork2014,dpcomposition2007} to help us build a complex workflow that still has the DP guarantee: sequential composition and postprocessing. Let $M_1(\cdot)$ and $M_2(\cdot)$ be $\epsilon_1-$ and $\epsilon_2-$ differentially private algorithms, sequential composition guarantees that releasing the outputs of $M_1(D)$ and $M_2(D)$ satisfies $(\epsilon_1 +\epsilon_2)-$ DP. Postprocessing an output of a DP algorithm does not incur any additional loss of privacy. For example, releasing $M_1(D)$ and $M_2(M_1(D))$ still satisfies $\epsilon_1-$ DP.

\subsection{ROC Curve and AUC}

 In a binary classification problem, given a threshold $\theta$, a predicted score $s_i$ is predicted to be $1$ if $s_i \ge \theta$. Given the ground-truth label and the predicted label (at a given threshold $\theta$), we can quantify the accuracy of the classifier on the dataset with True positives (TP($\theta$)), False positives (FP($\theta$)), False negatives (FN($\theta$)), and True negatives (TN($\theta$)).
 
 \begin{itemize}
    \item True positives, TP($\theta$), are the data points in the test whose true label and predicted label equal $1$. \textit{i.e.} $y_i = 1$ and $s_i \ge \theta$
    \item False positives, FP($\theta$), are the data points in test whose true label is $0$ but the predicted label is $1$. \textit{i.e.} $y_i = 0$ and $s_i \ge \theta$.
  \item  False negatives, FN($\theta$), are data points whose true label is $1$ but the predicted label is $0$. \textit{i.e.} $y_i = 1$ and $s_i < \theta$. 
  \item True negatives, TN($\theta$), are data points whose true label is $0$ and the predicted label is $0$. \textit{i.e.} $y_i = 0$ and $s_i < \theta$. 
\end{itemize}

 The area under the receiver operating characteristic (ROC) curves plots TPR (x-axis) vs. FPR (y-axis) over all possible thresholds $\theta$, and AUC is the area under the ROC curve. True Positive Rate (TPR)  (i.e. recall) is defined as $TPR(\theta) = \frac{TP(\theta)}{TP(\theta) + FN(\theta)}$ and False Positive Rate (FPR) is defined as $FPR(\theta) = \frac{FP(\theta)}{FP(\theta) + TN(\theta)}$. If the classifier is good, the ROC curve will be close to the left and upper boundary and AUC will be close to $1.0$ (a perfect classifier). On the other hand, if the classifier is poor, the ROC curve will be close to  the line from $(0,0)$ to $(1,1)$ with AUC around $0.5$ (random prediction).

\subsection{Privacy Leakage in AUC.} 

Researchers have shown AUC computation can cause privacy leakage. Matthews and Harel~\cite{Roc2013} demonstrate that by using a subset of the ground-truth data and the computed ROC curve, the data underlying the ROC curve can be reproduced accurately. Stoddard el al.~\cite{Dp4ML2014} show that an attacker can determine the unknown label by simply enumerating over all labels, guessing the labels, and then checking  which guesses lead to the given ROC curve. They propose a differentially private ROC curve computation algorithm. 
They first privately choose a set of thresholds (with privacy budget $\epsilon_1$). By modeling TP and FP values as one-sided range queries, they can compute noisy TPRs and FPRs values (using privacy budget $\epsilon_2$). They also leveraged a postprocessing step to enforce the monotonicity of TPRs and FPRs.
However, the above method is not designed for FL settings. In this paper, we aim to provide a differentially private way to compute AUC in the FL setting. 

\section{Threat Model}
\label{sec:threat}

In our hFL setting, there are multiple label parties (i.e. clients/devices) that own private labels (i.e. Y) and there is a central non-label party (i.e. server) that  is responsible for computing global AUC from all clients. 
The model is trained using the normal hFL protocol.


Our work focuses on the evaluation time and the goal of the server is to compute global AUC without letting clients directly share their private test data. In other words, clients cannot directly send the test data (i.e. private labels and prediction scores) to the server for it to compute AUC. Specifically, we are interested in protecting label information and therefore it is required that the exchanged information between client and server excludes the ground-truth test labels (Y) and corresponding prediction scores.

\section{Methods}
\label{sec:method_labeldp}

\begin{figure*}[ht!]
  \centering
  \includegraphics[width=1.0\linewidth]{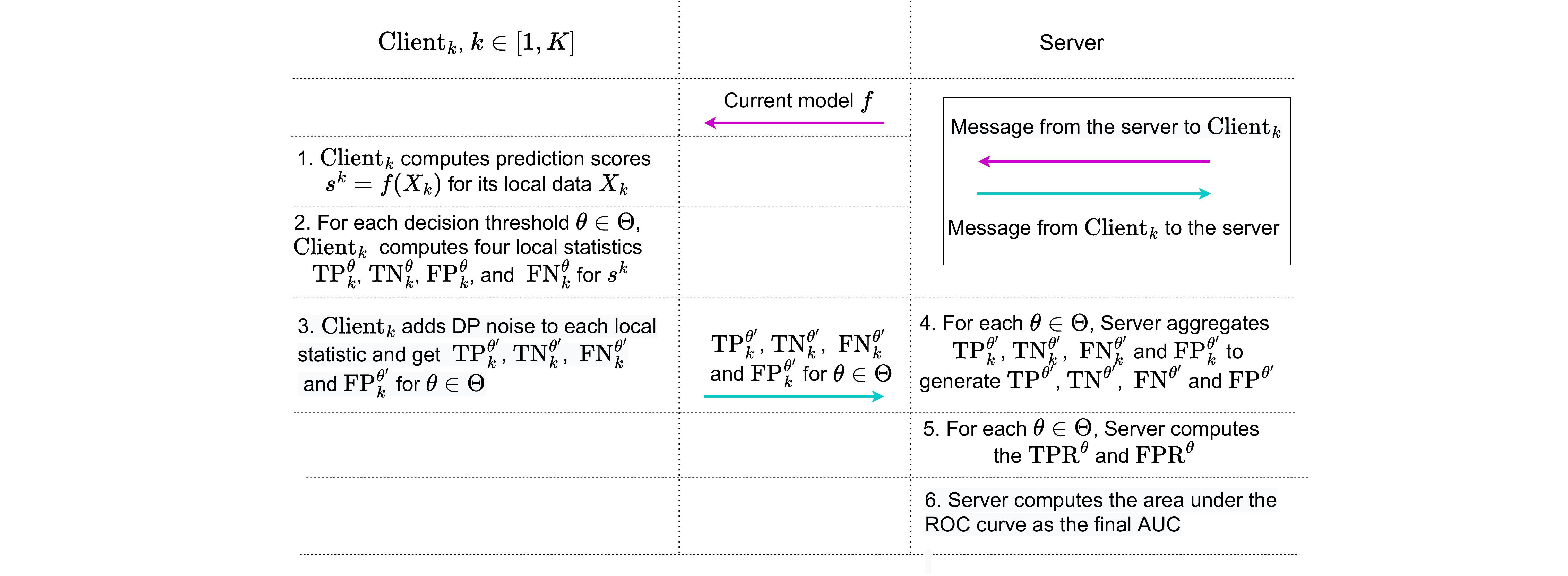}
  \caption{Illustration of our proposed \ourapp{} with  Laplace mechanism as an example.}
\label{fig:fedauc_illustration_lap_wo_sharing_prediction} 
 \end{figure*}

In this section, we introduce how to compute the AUC with label differential privacy by leveraging the Laplace mechanism. Here we use the Laplace mechanism as an example. Settings for other DP mechanisms such as the Gaussian mechanism will be the same. 

\subsection{Overall Workflow}
The workflow of this method is shown in Figure ~\ref{fig:fedauc_illustration_lap_wo_sharing_prediction}. The algorithm  has six steps:

\begin{enumerate}[leftmargin=*]
    \item \textbf{Clients Execute.} Each client $C_k$ computes the prediction scores $s_k = f(X_k)$ for all its owning data points.  
    \item \textbf{Clients Execute.} For each decision threshold $\theta \in \Theta$, client $C_k$ computes four local statistics $\text{TP}_k^\theta$,$\text{TN}_k^\theta$,$\text{FP}_k^\theta$, and $\text{FN}_k^\theta$ given the prediction scores $s_k$. 
    \item \textbf{Clients Executes.} The client $C_k$ adds perturbation with DP guarantee to  each local statistics $\text{TP}_k^\theta$,$\text{TN}_k^\theta$,$\text{FP}_k^\theta$, and $\text{FN}_k^\theta$ and get corresponding noisy statistics $\text{TP}_k^{\theta '}$,$\text{TN}_k^{\theta '}$,$\text{FP}_k^{\theta '}$, and $\text{FN}_k^{\theta '}$ for each $\theta \in \Theta$. Client $C_k$ sends all noisy statistics to the server.
    \item \textbf{Server Executes.} For each  $\theta \in \Theta$, the server aggregates the noisy statistics from all the clients: $\text{TP}^{\theta} = \sum_{k=1}^K \text{TP}_k^{\theta '}$,$\text{TN}^{\theta} = \sum_{k=1}^K \text{TN}_k^{\theta '}$,$\text{FP}^{\theta} = \sum_{k=1}^K \text{FP}_k^{\theta '}$, and $\text{FN}^{\theta} = \sum_{k=1}^K \text{FN}_k^{\theta '}$ 
    \item \textbf{Server Executes.} For each $\theta \in \Theta$, the server computes the corresponding  $\text{TPR}^\theta = \frac{\text{TP}^\theta}{\text{TP}^\theta + \text{FN}^\theta}$ and  $\text{FPR}^\theta = \frac{\text{FP}^\theta}{\text{FP}^\theta + \text{TN}^\theta}$.
    \item \textbf{Server Executes.}  The server plots TPR (x-axis) vs. FPR (y-axis) over all possible thresholds $\theta$ and computes the area under the corresponding  curve as AUC. 
\end{enumerate}

\subsection{Adding DP Noise to Local Statistics}
\label{sec:adding_noise_to_local_statistics}

In this section, we explain in details on how to perturb $\text{TP}$, $\text{TN}$,$\text{FP}$, and $\text{FN}$ for each $\theta \in \Theta$ in each client. It's worth mentioning that both Gaussian and Laplace mechanisms can be leveraged to generate the corresponding DP noise. Without loss of generality, we use Laplace as an example. Laplace mechanism preserves $(\epsilon, 0)$-differential privacy if the random noise 
 is drawn from the Laplace distribution $\textsf{Lap}(\frac{\Delta}{\epsilon})$ with parameter $\Delta/\epsilon$
 where $\Delta$ is the $l_1$ sensitivity ~\cite{dpdwork2014} and $\epsilon$ is the corresponding privacy budget. We name this method as \ourapplap{}. The noise is added as the following:

\begin{enumerate}[leftmargin=*]
    \item Adding noise to $\text{TP}$: Each client $C_k$ sets the corresponding sensitivity $\Delta_{\text{TP}_k^\theta} = 1$ for each $\theta \in \Theta$. Given a privacy budget $\epsilon_{\text{TP}}$, client $C_k$ draws the random noise from $\textsf{Lap}(1/\epsilon_{\text{TP}})$ and add it to $\text{TP}_k^\theta$ and get $\text{TP}_k^{\theta '}$ .
    \item Adding noise to $\text{FP}$: Each client $C_k$ sets the corresponding sensitivity $\Delta_{\text{FP}_k^\theta} = 1$ for each $\theta \in \Theta$. Given a privacy budget $\epsilon_{\text{FP}}$, client $C_k$ draws the random noise from $\textsf{Lap}(1/\epsilon_{\text{FP}})$ and add it to $\text{FP}_k^\theta$ and get $\text{FP}_k^{\theta '}$ .
    \item Adding noise to $\text{TN}$: Each client $C_k$ sets the corresponding sensitivity $\Delta_{\text{TN}_k^\theta} = 1$ for each $\theta \in \Theta$. Given a privacy budget $\epsilon_{\text{TN}}$, client $C_k$ draws the random noise from $\textsf{Lap}(1/\epsilon_{\text{TN}})$ and add it to $\text{TN}_k^\theta$ and get $\text{TN}_k^{\theta '}$ .
    \item Adding noise to $\text{FN}$: Each client $C_k$ sets the corresponding sensitivity $\Delta_{\text{FN}_k^\theta} = 1$ for each $\theta \in \Theta$. Given a privacy budget $\epsilon_{\text{FN}}$, client $C_k$ draws the random noise from $\textsf{Lap}(1/\epsilon_{\text{FN}})$ and add it to $\text{FN}_k^\theta$ and get $\text{FN}_k^{\theta '}$ .
\end{enumerate}

\para{Privacy Analysis.}
Based on the Composition Theorem of DP ~\cite{dpdwork2014}, the  privacy budget for each decision boundary ($\theta$) is ($\epsilon_{\text{TP}} + \epsilon_{TN} + \epsilon_{\text{FP}} + \epsilon_{FN}$). Since we have $|\Theta|$ decision thresholds ($\theta$), the total DP privacy budget is $\epsilon = |\Theta|*(\epsilon_{\text{TP}} + \epsilon_{TN} + \epsilon_{\text{FP}} + \epsilon_{FN})$. Without loss of generality, we set $\epsilon_{\text{TP}} = \epsilon_{TN} = \epsilon_{\text{FP}} = \epsilon_{FN}=\epsilon'$ in our paper and the total DP budget is $4|\Theta|\epsilon'$.

\section{Experiments}
\label{sec:experiments}

In this section, we show the experimental results of evaluating our proposed approaches. We introduce the experimental setups first. 

\begin{table*}[!htp]\centering
\scriptsize 
\setlength{\tabcolsep}{0.5em} 
{\renewcommand{\arraystretch}{1.2}
\begin{tabular}{c|c|c|c|c|c|c}\toprule
& & &\multicolumn{4}{c}{RR} \\ \hline 
epoch &Tensorflow &scikit-learn &$\epsilon$=8.0 &$\epsilon$=4.0 &$\epsilon$=2.0 &$\epsilon$=1.0 \\ \hline 
0 &0.749357 &0.749383 &0.74938 $\pm$ 0.00004 &0.749398 $\pm$ 0.0004 &0.749108 $\pm$ 0.000932 &0.750239 $\pm$ 0.001766 \\
1 &0.766434 &0.766477 &0.766447 $\pm$ 0.000054 &0.766338 $\pm$ 0.000338 &0.7665588 $\pm$ 0.001109 &0.766392 $\pm$ 0.002098 \\
2 &0.770189 &0.770219 &0.770204 $\pm$ 0.000049 &0.770112 $\pm$ 0.000375 &0.7702048 $\pm$ 0.000795 &0.771098 $\pm$ 0.001891 \\
\bottomrule
\end{tabular}}
\caption{ \ourapprr{} with $|\Theta| = 200$ and no noise added to the prediction scores}\label{tab:rr_roc_auc_thresholds_200}
\end{table*}

\subsection{Experimental Setup}

\textbf{Dataset.}
We evaluate the proposed approaches on Criteo~\footnote{{https://www.kaggle.com/c/criteo-display-ad-challenge/data}}, which is a large-scale industrial binary classification dataset (with approximately $45$ million user click records) for conversion prediction tasks. Every record of Criteo has $27$ categorical input features and $14$ real-valued input features. We first replace all the \texttt{NA} values in categorical features with a single new category (which we represent using the empty string) and replace all the \texttt{NA} values in real-valued features with $0$. For each categorical feature, we convert each of its possible value uniquely to an integer between $0$ (inclusive) and the total number of unique categories (exclusive). For each real-valued feature, we linearly normalize it into $[0,1]$. We then randomly sample $90\%$ of the entire Criteo set as our training data and the remaining $10\%$ as our test data. 
We computed the AUC on the test set which contains $M=458,407$ where $P = 117,317$ and $N = 341,090$ for $3$ epochs.

\textbf{Model.}
 We modified a popular deep learning model architecture WDL \cite{cheng2016wide} for online advertising. 
 Note that our goal is not to train the model that can beat the state-of-the-art, but to test the effectiveness of our proposed federated AUC computation approach. 

\textbf{Ground-truth AUC.}
We compare our proposed \ourapp{} with two AUC computation libraries (their computed results work as ground-truth and have no privacy guarantee): 1) scikit-learn\footnote{\url{https://scikit-learn.org/stable/modules/generated/sklearn.metrics.auc.html}}; 2) Tensorflow \footnote{\url{https://www.tensorflow.org/api_docs/python/tf/keras/metrics/AUC}}. Both approximate the AUC (Area under the curve) of the ROC. In our experiments, we set $\text{num\_thresholds}=1,000$ for Tensorflow. We use the default values for other parameters.

\textbf{Evaluation Metric.} For each method, we run the same setting for $100$ times (change the random seed every time) and use the corresponding mean and standard deviation of the computed AUC as our evaluation metric. 
A good computation method should achieve a small std of the computed AUC and the corresponding mean value of the computed AUC should be close to the ground-truth AUC. 



\subsection{One Randomized Responses based Competitor: \ourapprr{}}


\begin{figure*}[ht!]
  \centering
  \includegraphics[width=1.0\linewidth]{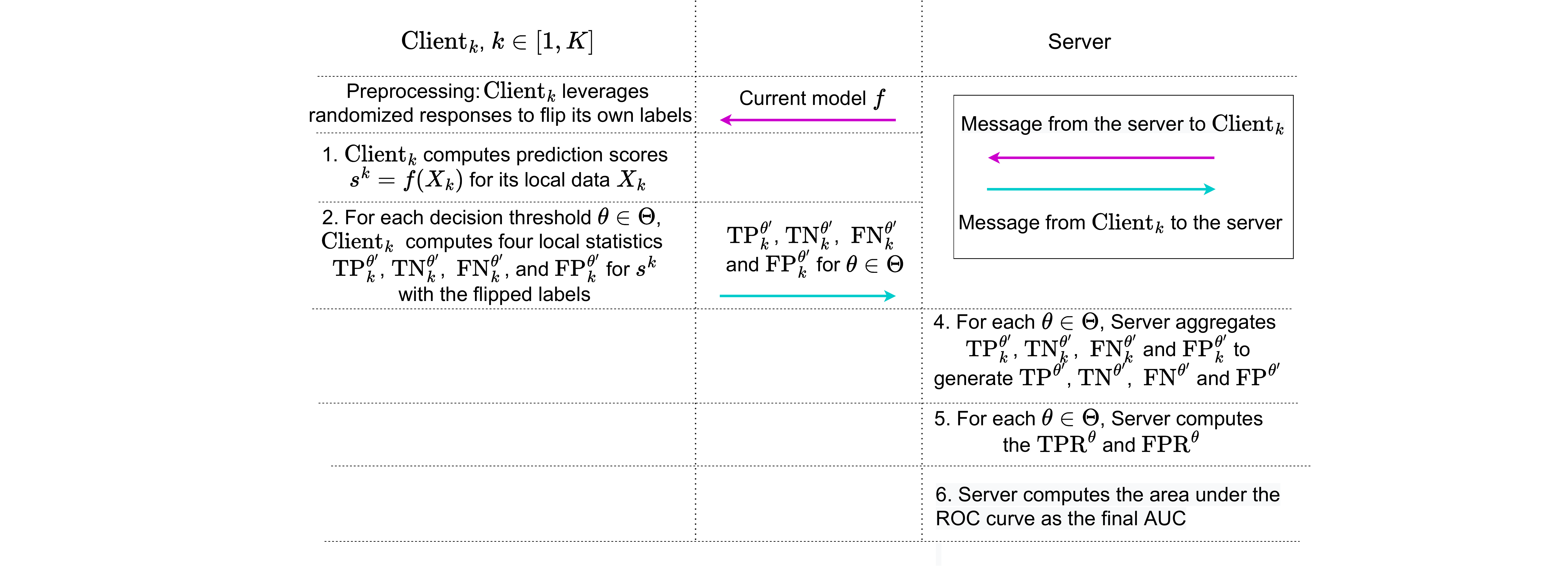}
  \caption{Illustration of \ourapprr{} with Randomized Responses mechanism.}
\label{fig:fedauc_illustration_rr_wo_sharing_prediction} 
 \end{figure*}
 We introduce another algorithm to compute AUC for evaluating FL models as one competitor. The algorithm is based on randomized response~\cite{RR} and we name it as \ourapprr{}. We include the workflow in Figure ~\ref{fig:fedauc_illustration_rr_wo_sharing_prediction}. We now explain the algorithm step by step.


\subsubsection{Step 1: Clients Flip Their Local Labels}

Randomized response (RR) is $\epsilon$-LabelDP ~\cite{localdp_rr2021,localdp2020} and  works as follows: let $\epsilon$ be a parameter and let $y \in \{0, 1\}$ be the true label. Given a query of $y$, RR will respond with a random draw $\tilde{y}$ from the following probability distribution: 


\begin{equation}
\label{eq:rr}
    \Pr[\tilde y = \hat y] = \begin{cases}
     \frac{e^\epsilon}{1+e^\epsilon} & \text{for $y = \hat y$,} \\
    \frac{1}{1+e^\epsilon} & \text{otherwise.}
                            \end{cases}
\end{equation}




Clients leverage randomized responses to flip their owning labels as a preprocessing step before computing the AUC. It's worth mentioning that all labels are only flipped once and the generated noisy labels can then be used for further evaluations multi-times. 


\subsubsection{Step 2: Server Computes AUC from Flipped Labels}
\label{sec:corr_auc}

 Since the corresponding AUC is computed with flipped labels,
 we denote this AUC as \emph{noisy AUC}: $\text{AUC}^{\text{noisy}}$. It has the same six steps as \ourapp{} except that each client computes local statistics with flipped labels and sends the corresponding results (without adding additional noises) to the server.

\subsubsection{Step 3: Server Debiases AUC}
\label{sec:covert_auc_to_real}

We leveraged \cite{ConvertAUC2015} to debiase the noisy AUC and get the final clean AUC $\text{AUC}^{{\text{clean}}}$ that we are interested in.

\subsubsection{Utility and Privacy Analysis}

The corresponding results of \ourapprr{} can be seen in Table ~\ref{tab:rr_roc_auc_thresholds_200}. However, \ourapprr{} has potential privacy issues since the prediction scores can be inferred from the change of adjacent local statistics. For example, we can conclude that there are some prediction scores that fall in $\theta_{i+1}$ if there are some differences between ($\text{TP}_k^{\theta_{i+1} '}$, $\text{TP}_k^{\theta_{i+1} '}$, $\text{TP}_k^{\theta_{i+1} '}$,  $\text{TP}_k^{\theta_{i+1} '}$) and ($\text{TP}_k^{\theta_{i} '}$, $\text{TP}_k^{\theta_{i} '}$, $\text{TP}_k^{\theta_{i} '}$, $\text{TP}_k^{\theta_{i} '}$) from client $k$.

\begin{figure}[ht!]
  \centering
    \includegraphics[width=0.75\linewidth]{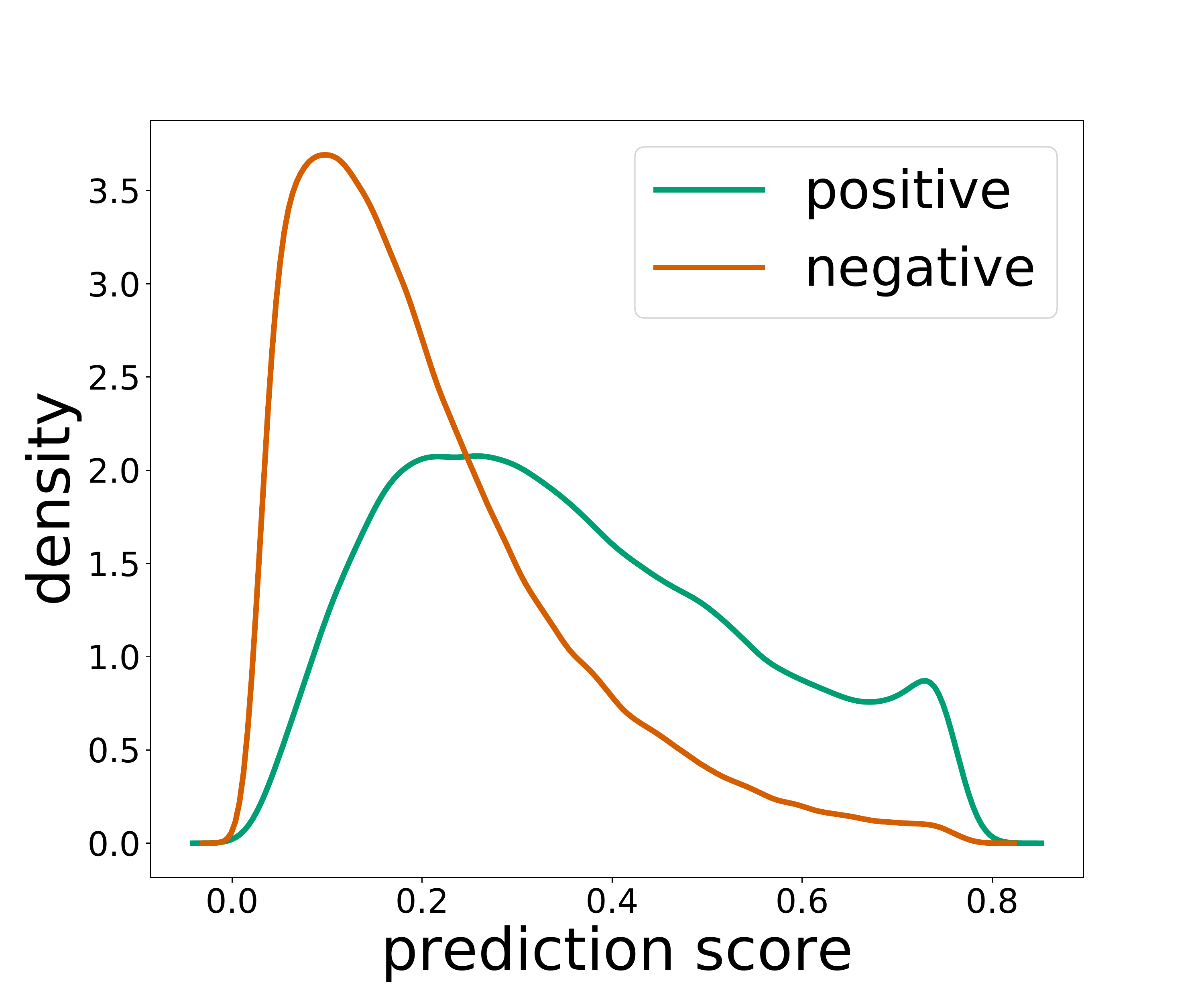}
  \caption{Density of positive and negative instances' prediction scores in Criteo data.}
 \label{fig:criteo_negative_positive_density} 
 \end{figure}

Attackers can then infer the label information based on the exposed prediction scores. We propose a simple attack method to infer the label information based on the prediction scores. The strategy of the attack is to select the samples with top-K prediction scores as positive labels. We measure the corresponding guessing performance by precision and recall. As shown in Figure ~\ref{fig:criteo_negative_positive_density} and Table ~\ref{tab:top_k_analysis_criteo_data_small}, positive instances can have relatively higher prediction scores than negative ones at some density areas. For example, we can achieve a $79\%$ precision if we select the instances with top-$100$ prediction scores. Hence, exposing prediction scores among all participants can increase the risk of label leakage.  However, since the prediction scores $s^k$ are shuffled before sending to the server, the server has no idea which prediction score belongs to which data sample \footnote{$|s^k| \geq 2$}. 

\begin{table}[!htp]\centering
\setlength{\tabcolsep}{0.5em} 
{\renewcommand{\arraystretch}{1.2}
\begin{tabular}{c|c|c|c}\toprule
Top K & $\#$ Positives in Top K &Precision &Recall \\ \hline
1 &1 &1 &8.52e-6 \\ \hline
5 &4 &0.8 &3.41e-5 \\ \hline
10 &8 &0.8 &6.82e-5 \\ \hline
50 &43 &0.86 &3.67e-4 \\ \hline
100 &79 &0.79 &6.73e-4 \\ \hline
500 &384 &0.768 &3.27e-3 \\ \hline
1,000 &774 &0.774 &6.60e-3 \\ \hline
5,000 &3,914 &0.7828 &0.0334 \\ \hline
10,000 &7,519 &0.7519 &0.0641 \\ \hline
50,000 &30,793 &0.6159 &0.2625 \\ \hline
100,000 &52,268 &0.5227 &0.4455 \\
\bottomrule
\end{tabular}}
\caption{Top-k analysis on Criteo data }\label{tab:top_k_analysis_criteo_data_small}
\end{table}

To prevent label leakage from the prediction scores, we may leverage DP to add noise to the prediction scores. Since the prediction score is the output of a softmax/sigmoid function, the corresponding sensitivity $\Delta = 1$. We can then leverage the Laplace mechanism \footnote{The Gaussian mechanism can be applied here too} to add corresponding noise to the prediction scores. The corresponding results can be seen in Table ~\ref{tab:laplace_dp_for_prediction_score_criteo}. We can observe that the utility of the computed AUC is highly sensitive to the privacy budget of the prediction scores. We cannot even achieve a reasonable AUC utility with a small $\epsilon$ (i.e. $\epsilon \le 10$) of prediction scores. The results show that \ourapprr{} is not a suitable procedure to achieve a good tradeoff between utility and privacy for computing AUC in FL. 

\begin{table}[ht!]\centering
\setlength{\tabcolsep}{0.5em} 
{\renewcommand{\arraystretch}{1.2}
\begin{tabular}{c|c|c|c|c}\toprule
&Epoch &0 &1 &2 \\ \hline
&Tensorflow &0.7494 &0.7665 &0.7702 \\ \hline 
&scikit-learn &0.7494 &0.7665 &0.7702 \\ \hline 
\multirow{13}{*}{$\epsilon$} &1 &0.5374 &0.5416 &0.5420 \\ \cline{2-5}
&2 &0.5725 &0.5795 &0.5830 \\ \cline{2-5}
&3 &0.5976 &0.6105 &0.6130 \\ \cline{2-5}
&4 &0.6239 &0.6367 &0.6387 \\ \cline{2-5}
&5 &0.6413 &0.6569 &0.6607 \\ \cline{2-5}
&6 &0.6571 &0.6732 &0.6786 \\ \cline{2-5}
&7 &0.6698 &0.6870 &0.6903 \\ \cline{2-5}
&8 &0.6797 &0.6976 &0.7022 \\ \cline{2-5}
&9 &0.6885 &0.7060 &0.7115 \\ \cline{2-5}
&10 &0.6952 &0.7124 &0.7191 \\ \cline{2-5}
&50 &0.7454 &0.7625 &0.7663 \\ \cline{2-5}
&100 &0.7484 &0.7654 &0.7692 \\ \cline{2-5}
&1000 &0.7494 &0.7665 &0.7702 \\ \cline{2-5}
\hline
\end{tabular}}
\caption{AUC calculated with $\epsilon$-DP for prediction scores with Laplace mechanism on Criteo dataset}\label{tab:laplace_dp_for_prediction_score_criteo}
\end{table}

\subsection{Experimental Results of \ourapplap{}}

We now introduce the experimental results of \ourapplap{} and demonstrate the effectiveness of \ourapplap{}.

\subsubsection{IID vs. Non-IID}

Based on the setting of how to assign data samples to clients, we provide two simulations to conduct the corresponding experiments.

\begin{enumerate}[leftmargin=*]
    \item IID: all data points are uniformly assigned to the clients. 
    \item Non-IID: all data points are assigned to the clients based on their corresponding prediction scores. Data samples with similar prediction scores will be assigned to the same client. 
\end{enumerate}

\begin{table*}[!ht]\centering
\scriptsize 
\setlength{\tabcolsep}{0.5em} 
{\renewcommand{\arraystretch}{1.2}
\begin{tabular}{c|c|c|c|c|c|c|cc}\toprule
epoch &Tensorflow &scikit-learn & &$\epsilon'$=0.02, $\epsilon$=8 &$\epsilon'$=0.01, $\epsilon$=4 &$\epsilon'$=0.005, $\epsilon$=2 &$\epsilon'$=0.0025,$\epsilon$=1 \\ \hline 
\multirow{4}{*}{0} &\multirow{4}{*}{0.749357} &\multirow{4}{*}{0.749383} &IID, $K=10$ &0.749347 $\pm$ 0.000216 &0.749178 $\pm$ 0.000484 &0.748937 $\pm$ 0.000755 &0.748206 $\pm$ 0.001649 \\ \cline{4-8}
& & &Non-IID, $K=10$ &0.749319 $\pm$ 0.000242 &0.749260 $\pm$ 0.000474 &0.749192 $\pm$ 0.00072 &0.748962 $\pm$ 0.00178 \\ \cline{4-8}
& & &IID, $K=1,000$ &0.74811 $\pm$ 0.002335 &0.745557 $\pm$ 0.004498 &0.738944 $\pm$ 0.008008 &0.726187 $\pm$ 0.014704 \\ \cline{4-8}
& & &Non-IID, $K=1,000$ &0.748062 $\pm$ 0.001981 &0.746105 $\pm$ 0.004465 &0.741711 $\pm$ 0.009246 &0.726885 $\pm$ 0.016539 \\ \hline 
\multirow{4}{*}{1} &\multirow{4}{*}{0.766434} &\multirow{4}{*}{0.766477} &IID, $K=10$ &0.766373 $\pm$ 0.000239 &0.766309 $\pm$ 0.000402 &0.76611 $\pm$ 0.000894 &0.765684 $\pm$ 0.001677 \\ \cline{4-8}
& & &Non-IID, $K=10$ &0.766378 $\pm$ 0.000231 &0.766197 $\pm$ 0.000471 &0.76612 $\pm$ 0.000755 &0.765447 $\pm$ 0.001805 \\ \cline{4-8}
& & &IID, $K=1,000$ &0.764887 $\pm$ 0.002241 &0.763012 $\pm$ 0.004378 &0.756938 $\pm$ 0.008096 &0.740619 $\pm$ 0.017457 \\ \cline{4-8}
& & &Non-IID, $K=1,000$ &0.764923 $\pm$ 0.002036 &0.763246 $\pm$ 0.004022 &0.756715 $\pm$ 0.009105 &0.744978 $\pm$ 0.018234 \\ \hline 
\multirow{4}{*}{2} &\multirow{4}{*}{0.770189} &\multirow{4}{*}{0.770219} &IID, $K=10$ &0.770091 $\pm$ 0.000258 &0.769995 $\pm$ 0.000498 &0.770028 $\pm$ 0.000788 &0.769143 $\pm$ 0.00172 \\ \cline{4-8}
& & &Non-IID, $K=10$ &0.770095 $\pm$ 0.000213 &0.770115 $\pm$ 0.000434 &0.76981 $\pm$ 0.000903 &0.769265 $\pm$ 0.001502 \\ \cline{4-8}
& & &IID, $K=1,000$ &0.76918 $\pm$ 0.002519 &0.766482 $\pm$ 0.004387 &0.761074 $\pm$ 0.007947 &0.745812 $\pm$ 0.014596 \\ \cline{4-8}
& & &Non-IID, $K=1,000$ &0.768894 $\pm$ 0.002104 &0.766819 $\pm$ 0.003802 &0.759916 $\pm$ 0.007251 &0.746815 $\pm$ 0.016511 \\
\bottomrule
\end{tabular}}
\caption{IID vs. Non-IID and $K=10$ vs. $K=1,000$ with $|\Theta|=100$}\label{tab:dpauc_thresholds_100_iidvsnoniid_varying_clients}
\end{table*}

We divide $M$ data samples into $K$ clients based on the IID and Non-IID settings and each client has $\frac{M}{K}$ data samples on average.  We set $|\Theta| = 100$ and compare the difference between IID and Non-IID settings for $K=10$ and $K=1,000$ in Table ~\ref{tab:dpauc_thresholds_100_iidvsnoniid_varying_clients}. We can observe that our method can achieve similar performance (mean and standard deviation of computed AUC) under both IID and Non-IID settings. It indicates that our proposed method is robust to Non-IID settings. 

\subsubsection{Effects of Number of Samples per Client}

Given the fixed total number of data points ($M$), \ourapplap{} is sensitive to the number of clients or the number of samples per client has. We vary the number of data samples per client owing and conduct experiments with $K=10$ (around $45,840$ data points per client) and $1,000$ (around $458$ data points per client). The corresponding results are shown in Table ~\ref{tab:dpauc_thresholds_100_iidvsnoniid_varying_clients}. We can conclude that  with an increasing avg. $\#$ data samples per client, \ourapplap{} can achieve a smaller standard deviation of AUC estimation since it adds relatively small amounts of noise to the local statistics and hence has fewer effects on the computational results.

\subsubsection{Effects of Number of Decision Boundaries}

We also conducted experiments to test the effects of the number of decision boundaries ($|\Theta|$). The corresponding experiments are performed with IID setting with $K=10$. We tested $|\Theta| = 10, 25, 50, 100, 200$ as shown in Table ~\ref{tab:dpauc_iid_clients_10_varying_thresholds}. Given the same privacy budget $\epsilon$, with increasing $|\Theta|$, each local statistic has to be assigned a smaller privacy budget $\epsilon'$ (adding more DP noise). As a result, the standard deviation of the computed AUC will be smaller for smaller $|\Theta|$. However, smaller $|\Theta|$ can have a worse effect on the computed precision of the resulted AUC. In our experiments, we found $|\Theta|=100$ can achieve a good performance on both precision and standard deviation of the AUC.

\begin{table*}[!ht]\centering
\scriptsize
\setlength{\tabcolsep}{0.5em} 
{\renewcommand{\arraystretch}{1.2}
\begin{tabular}{c|c|c|c|c|c|c|cc}\toprule
epoch &Tensorflow &scikit-learn &$|\Theta|$ & $\epsilon=8$ &$\epsilon=4$ &$\epsilon=2$ &$\epsilon=1$ \\ \hline 
\multirow{5}{*}{0} &\multirow{5}{*}{0.749357} &\multirow{5}{*}{0.749383} &10 &0.7251052 $\pm$ 0.000074 &0.725064 $\pm$ 0.000167 &0.725008 $\pm$ 0.000323 &0.724899 $\pm$ 0.000607 \\ \cline{4-8}
& & &25 &0.747879 $\pm$ 0.000147 &0.747895 $\pm$ 0.000275 &0.747669 $\pm$ 0.000625 &0.747278 $\pm$ 0.000889 \\ \cline{4-8}
& & &50 &0.748914 $\pm$ 0.000161 &0.748884 $\pm$ 0.000343 &0.748988 $\pm$ 0.000657 &0.748816 $\pm$ 0.001279 \\ \cline{4-8}
& & &100 &0.749347 $\pm$ 0.000216 &0.749178 $\pm$ 0.000484 &0.748937 $\pm$ 0.000755 &0.748206 $\pm$ 0.001649 \\ \cline{4-8}
& & &200 &0.74929 $\pm$ 0.000289 &0.74916 $\pm$ 0.000573 &0.748491 $\pm$ 0.001204 &0.747117 $\pm$ 0.002224 \\ \hline 
\multirow{5}{*}{1} &\multirow{5}{*}{0.766434} &\multirow{5}{*}{0.766477} &10 &0.744348 $\pm$ 0.000084 &0.744317 $\pm$ 0.000144 &0.744215 $\pm$ 0.000328 &0.744058 $\pm$ 0.000639 \\ \cline{4-8}
& & &25 &0.765178 $\pm$ 0.000159 &0.765045 $\pm$ 0.000252 &0.7649144 $\pm$ 0.000507 &0.764968 $\pm$ 0.001282 \\ \cline{4-8}
& & &50 &0.766198 $\pm$ 0.000187 &0.766181 $\pm$ 0.000329 &0.765973 $\pm$ 0.000723 &0.766009 $\pm$ 0.001216 \\ \cline{4-8}
& & &100 &0.766373 $\pm$ 0.000239 &0.766309 $\pm$ 0.000402 &0.76611 $\pm$ 0.000894 &0.765684 $\pm$ 0.001677 \\ \cline{4-8}
& & &200 &0.76638 $\pm$ 0.000311 &0.766235 $\pm$ 0.000677 &0.765694 $\pm$ 0.001367 &0.764399 $\pm$ 0.002264 \\ \hline 
\multirow{5}{*}{2} &\multirow{5}{*}{0.770189} &\multirow{5}{*}{0.770219} &10 &0.748858 $\pm$ 0.000073 &0.7487984 $\pm$ 0.000123 &0.748821 $\pm$ 0.000331 &0.748626 $\pm$ 0.000591 \\ \cline{4-8}
& & &25 &0.769006 $\pm$ 0.000162 &0.768992 $\pm$ 0.000323 &0.768705 $\pm$ 0.000508 &0.768357 $\pm$ 0.001308 \\ \cline{4-8}
& & &50 &0.769806 $\pm$ 0.000220 &0.769756 $\pm$ 0.000500 &0.769549 $\pm$ 0.000805 &0.76939 $\pm$ 0.001034 \\ \cline{4-8}
& & &100 &0.770091 $\pm$ 0.000258 &0.769995 $\pm$ 0.000498 &0.770028 $\pm$ 0.000788 &0.769143 $\pm$ 0.00172 \\ \cline{4-8}
& & &200 &0.770137 $\pm$ 0.000319 &0.769729 $\pm$ 0.000633 &0.769066 $\pm$ 0.001085 &0.767694 $\pm$ 0.002746 \\
\bottomrule
\end{tabular}}
\caption{Sensitivity of number of thresholds ($|\Theta|$) with IID setting and $K=10$.}\label{tab:dpauc_iid_clients_10_varying_thresholds}
\end{table*}

\section{Related Work}
\label{sec:related_work}

\para{Federated Learning.} FL~\cite{mmr+17,yang2019federated} can be mainly classified into three categories: \textit{horizontal FL},  \textit{vFL}, and \textit{federated transfer learning} ~\cite{yang2019federated}. In this paper, we focus on hFL which partitions data by entities.  

\textbf{Information Leakage in FL.} Recently, studies show that in FL, even though the raw data (feature and label) is not shared,
sensitive information can still be leaked from the gradients and intermediate embeddings communicated between parties. For example, \cite{vepakomma2019reducing} and \cite{sunDefending2021} showed that server's raw features can be leaked from the forward cut layer embedding.  In addition, \cite{li2021label} studied the label leakage problem but the leakage source was the backward gradients rather than forward embeddings.  ~\cite{zhu2019dlg} showed that an honest-but-curious server can uncover the raw features and labels of a device by knowing the model architecture, parameters, and  communicated gradient of the loss on the device's data. Based on their techniques, \cite{zhao2020idlg} showed that the ground truth label of an example can be extracted by exploiting 
the directions of the gradients of the weights connected to the logits of different classes. 

\textbf{Information Protection in FL.} 
There are three main categories of information protection techniques in FL:
\textbf{1)} cryptographic methods such as {secure multi-party computation} \cite{bonawitz2017practical}; \textbf{2)} system-based methods including {trusted execution environments} \cite{subramanyan2017formal}; and \textbf{3)} perturbation methods that add noise to the communicated messages~\cite{abadi2016deep, mcmahan2017learning,erlingsson2019amplification,cheu2019distributed,zhu2019dlg}. In this paper, we focus on adding DP noise to protect the private label information during computing AUC in FL.
\section{Conclusion}
\label{sec:conclusions}

In this paper, we focus on providing label differential privacy for computing AUC during the model evaluation in the setting of hFL. We proposed an approach with label DP to compute AUC for evaluating models. We conducted extensive experiments to verify the effectiveness of our proposed methods. We use the Laplace mechanism as an example. Other DP mechanisms such as the Gaussian mechanism can be applied to our framework too. In our current design, the privacy budget of \ourapplap is linear with the query times (model evaluation times on the same evaluation set). As future work, we may leverage the strong composition property of DP ~\cite{boostingdp2010,numericaldp2021} to reduce the privacy budget from linear to square root.  




\clearpage
\bibliography{references}

\begin{thebibliography}{46}
\providecommand{\natexlab}[1]{#1}

\bibitem[{Abadi et~al.(2016)Abadi, Chu, Goodfellow, McMahan, Mironov, Talwar,
  and Zhang}]{abadi2016deep}
Abadi, M.; Chu, A.; Goodfellow, I.; McMahan, H.~B.; Mironov, I.; Talwar, K.;
  and Zhang, L. 2016.
\newblock Deep learning with differential privacy.
\newblock In \emph{Proceedings of the 2016 ACM SIGSAC Conference on Computer
  and Communications Security}, 308--318.

\bibitem[{Abuadbba et~al.(2020)Abuadbba, Kim, Kim, Thapa, Camtepe, Gao, Kim,
  and Nepal}]{akk+20}
Abuadbba, S.; Kim, K.; Kim, M.; Thapa, C.; Camtepe, S.~A.; Gao, Y.; Kim, H.;
  and Nepal, S. 2020.
\newblock Can We Use Split Learning on 1D CNN Models for Privacy Preserving
  Training?
\newblock In \emph{Proceedings of the 15th ACM Asia Conference on Computer and
  Communications Security}, 305--318.

\bibitem[{Bebensee(2019)}]{localdp2019}
Bebensee, B. 2019.
\newblock Local Differential Privacy: a tutorial.
\newblock \emph{CoRR}, abs/1907.11908.

\bibitem[{Bonawitz et~al.(2017)Bonawitz, Ivanov, Kreuter, Marcedone, McMahan,
  Patel, Ramage, Segal, and Seth}]{bonawitz2017practical}
Bonawitz, K.; Ivanov, V.; Kreuter, B.; Marcedone, A.; McMahan, H.~B.; Patel,
  S.; Ramage, D.; Segal, A.; and Seth, K. 2017.
\newblock Practical secure aggregation for privacy-preserving machine learning.
\newblock In \emph{Proceedings of the 2017 ACM SIGSAC Conference on Computer
  and Communications Security}, 1175--1191.

\bibitem[{Ceballos et~al.(2020)Ceballos, Sharma, Mugica, Singh, Roman,
  Vepakomma, and Raskar}]{csm+20}
Ceballos, I.; Sharma, V.; Mugica, E.; Singh, A.; Roman, A.; Vepakomma, P.; and
  Raskar, R. 2020.
\newblock SplitNN-driven Vertical Partitioning.
\newblock \emph{arXiv preprint arXiv:2008.04137}.

\bibitem[{Chen et~al.(2016)Chen, Machanavajjhala, Reiter, and
  Barrientos}]{DPRegressionDiag2016}
Chen, Y.; Machanavajjhala, A.; Reiter, J.~P.; and Barrientos, A.~F. 2016.
\newblock Differentially Private Regression Diagnostics.
\newblock In \emph{2016 IEEE 16th International Conference on Data Mining
  (ICDM)}, 81--90.

\bibitem[{Cheng et~al.(2016)Cheng, Koc, Harmsen, Shaked, Chandra, Aradhye,
  Anderson, Corrado, Chai, Ispir et~al.}]{cheng2016wide}
Cheng, H.-T.; Koc, L.; Harmsen, J.; Shaked, T.; Chandra, T.; Aradhye, H.;
  Anderson, G.; Corrado, G.; Chai, W.; Ispir, M.; et~al. 2016.
\newblock Wide \& deep learning for recommender systems.
\newblock In \emph{Proceedings of the 1st workshop on deep learning for
  recommender systems}, 7--10.

\bibitem[{Cheu et~al.(2019)Cheu, Smith, Ullman, Zeber, and
  Zhilyaev}]{cheu2019distributed}
Cheu, A.; Smith, A.; Ullman, J.; Zeber, D.; and Zhilyaev, M. 2019.
\newblock Distributed differential privacy via shuffling.
\newblock In \emph{Annual International Conference on the Theory and
  Applications of Cryptographic Techniques}, 375--403. Springer.

\bibitem[{Duchi, Jordan, and Wainwright(2013)}]{localdpDuchi2013}
Duchi, J.~C.; Jordan, M.~I.; and Wainwright, M.~J. 2013.
\newblock Local Privacy and Statistical Minimax Rates.
\newblock In \emph{2013 IEEE 54th Annual Symposium on Foundations of Computer
  Science}, 429--438.

\bibitem[{Dwork et~al.(2006)Dwork, McSherry, Nissim, and Smith}]{dmns06}
Dwork, C.; McSherry, F.; Nissim, K.; and Smith, A. 2006.
\newblock Calibrating noise to sensitivity in private data analysis.
\newblock In \emph{Theory of cryptography conference}, 265--284. Springer.

\bibitem[{Dwork and Roth(2014{\natexlab{a}})}]{dr14}
Dwork, C.; and Roth, A. 2014{\natexlab{a}}.
\newblock The algorithmic foundations of differential privacy.
\newblock \emph{Found. Trends Theor. Comput. Sci.}, 9(3-4): 211--407.

\bibitem[{Dwork and Roth(2014{\natexlab{b}})}]{dpdwork2014}
Dwork, C.; and Roth, A. 2014{\natexlab{b}}.
\newblock The Algorithmic Foundations of Differential Privacy.
\newblock \emph{Found. Trends Theor. Comput. Sci.}, 9(3–4): 211–407.

\bibitem[{Dwork, Rothblum, and Vadhan(2010)}]{boostingdp2010}
Dwork, C.; Rothblum, G.~N.; and Vadhan, S. 2010.
\newblock Boosting and Differential Privacy.
\newblock In \emph{2010 IEEE 51st Annual Symposium on Foundations of Computer
  Science}, 51--60.

\bibitem[{Erlingsson et~al.(2019)Erlingsson, Feldman, Mironov, Raghunathan,
  Talwar, and Thakurta}]{erlingsson2019amplification}
Erlingsson, {\'U}.; Feldman, V.; Mironov, I.; Raghunathan, A.; Talwar, K.; and
  Thakurta, A. 2019.
\newblock Amplification by shuffling: From local to central differential
  privacy via anonymity.
\newblock In \emph{Proceedings of the Thirtieth Annual ACM-SIAM Symposium on
  Discrete Algorithms}, 2468--2479. SIAM.

\bibitem[{Erlingsson, Pihur, and Korolova(2014)}]{localdprappor2014}
Erlingsson, {\'{U}}.; Pihur, V.; and Korolova, A. 2014.
\newblock {RAPPOR:} Randomized Aggregatable Privacy-Preserving Ordinal
  Response.
\newblock In Ahn, G.; Yung, M.; and Li, N., eds., \emph{Proceedings of the 2014
  {ACM} {SIGSAC} Conference on Computer and Communications Security,
  Scottsdale, AZ, USA, November 3-7, 2014}, 1054--1067. {ACM}.

\bibitem[{Geiping et~al.(2020)Geiping, Bauermeister, Dr\"{o}ge, and
  Moeller}]{Jonas2020}
Geiping, J.; Bauermeister, H.; Dr\"{o}ge, H.; and Moeller, M. 2020.
\newblock Inverting Gradients - How easy is it to break privacy in federated
  learning?
\newblock In Larochelle, H.; Ranzato, M.; Hadsell, R.; Balcan, M.~F.; and Lin,
  H., eds., \emph{Advances in Neural Information Processing Systems},
  volume~33, 16937--16947. Curran Associates, Inc.

\bibitem[{Ghazi et~al.(2021)Ghazi, Golowich, Kumar, Manurangsi, and
  Zhang}]{labeldp21}
Ghazi, B.; Golowich, N.; Kumar, R.; Manurangsi, P.; and Zhang, C. 2021.
\newblock On Deep Learning with Label Differential Privacy.
\newblock \emph{arXiv preprint arXiv:2102.06062}.

\bibitem[{Ghosh et~al.(2020)Ghosh, Chung, Yin, and Ramchandran}]{gcyr20}
Ghosh, A.; Chung, J.; Yin, D.; and Ramchandran, K. 2020.
\newblock An Efficient Framework for Clustered Federated Learning.
\newblock In Larochelle, H.; Ranzato, M.; Hadsell, R.; Balcan, M.~F.; and Lin,
  H., eds., \emph{Advances in Neural Information Processing Systems},
  volume~33, 19586--19597. Curran Associates, Inc.

\bibitem[{Gopi, Lee, and Wutschitz(2021)}]{numericaldp2021}
Gopi, S.; Lee, Y.~T.; and Wutschitz, L. 2021.
\newblock Numerical Composition of Differential Privacy.
\newblock In Ranzato, M.; Beygelzimer, A.; Dauphin, Y.; Liang, P.; and Vaughan,
  J.~W., eds., \emph{Advances in Neural Information Processing Systems},
  volume~34, 11631--11642. Curran Associates, Inc.

\bibitem[{Gupta and Raskar(2018)}]{gr18}
Gupta, O.; and Raskar, R. 2018.
\newblock Distributed learning of deep neural network over multiple agents.
\newblock \emph{Journal of Network and Computer Applications}, 116: 1--8.

\bibitem[{Hamer, Mohri, and Suresh(2020)}]{Jenny2020}
Hamer, J.; Mohri, M.; and Suresh, A.~T. 2020.
\newblock {F}ed{B}oost: A Communication-Efficient Algorithm for Federated
  Learning.
\newblock In III, H.~D.; and Singh, A., eds., \emph{Proceedings of the 37th
  International Conference on Machine Learning}, volume 119 of
  \emph{Proceedings of Machine Learning Research}, 3973--3983. PMLR.

\bibitem[{Hanzely et~al.(2020)Hanzely, Hanzely, Horv\'{a}th, and
  Richtarik}]{hhh+20}
Hanzely, F.; Hanzely, S.; Horv\'{a}th, S.; and Richtarik, P. 2020.
\newblock Lower Bounds and Optimal Algorithms for Personalized Federated
  Learning.
\newblock In Larochelle, H.; Ranzato, M.; Hadsell, R.; Balcan, M.~F.; and Lin,
  H., eds., \emph{Advances in Neural Information Processing Systems},
  volume~33, 2304--2315. Curran Associates, Inc.

\bibitem[{Karimireddy et~al.(2020)Karimireddy, Kale, Mohri, Reddi, Stich, and
  Suresh}]{Sai2020}
Karimireddy, S.~P.; Kale, S.; Mohri, M.; Reddi, S.; Stich, S.; and Suresh,
  A.~T. 2020.
\newblock {SCAFFOLD}: Stochastic Controlled Averaging for Federated Learning.
\newblock In III, H.~D.; and Singh, A., eds., \emph{Proceedings of the 37th
  International Conference on Machine Learning}, volume 119 of
  \emph{Proceedings of Machine Learning Research}, 5132--5143. PMLR.

\bibitem[{Kasiviswanathan et~al.(2008)Kasiviswanathan, Lee, Nissim,
  Raskhodnikova, and Smith}]{localdp2008}
Kasiviswanathan, S.~P.; Lee, H.~K.; Nissim, K.; Raskhodnikova, S.; and Smith,
  A.~D. 2008.
\newblock What Can We Learn Privately?
\newblock \emph{CoRR}, abs/0803.0924.

\bibitem[{Li et~al.(2022)Li, Sun, Yang, Gao, Zhang, Xie, Smith, and
  Wang}]{li2021label}
Li, O.; Sun, J.; Yang, X.; Gao, W.; Zhang, H.; Xie, J.; Smith, V.; and Wang, C.
  2022.
\newblock Label Leakage and Protection in Two-party Split Learning.
\newblock In \emph{The Tenth International Conference on Learning
  Representations (ICLR)}.

\bibitem[{Li et~al.(2020)Li, Kovalev, Qian, and Richtarik}]{Li2020}
Li, Z.; Kovalev, D.; Qian, X.; and Richtarik, P. 2020.
\newblock Acceleration for Compressed Gradient Descent in Distributed and
  Federated Optimization.
\newblock In III, H.~D.; and Singh, A., eds., \emph{Proceedings of the 37th
  International Conference on Machine Learning}, volume 119 of
  \emph{Proceedings of Machine Learning Research}, 5895--5904. PMLR.

\bibitem[{Ma and Wang(2021)}]{localdp_rr2021}
Ma, F.; and Wang, P. 2021.
\newblock Randomized Response Mechanisms for Differential Privacy Data
  Analysis: Bounds and Applications.
\newblock \emph{CoRR}, abs/2112.07397.

\bibitem[{Matthews and Harel(2013{\natexlab{a}})}]{aucleak2013}
Matthews, G.; and Harel, O. 2013{\natexlab{a}}.
\newblock An Examination of Data Confidentiality and Disclosure Issues Related
  to Publication of Empirical ROC Curves.
\newblock \emph{Academic radiology}, 20: 889--96.

\bibitem[{Matthews and Harel(2013{\natexlab{b}})}]{Roc2013}
Matthews, G.~J.; and Harel, O. 2013{\natexlab{b}}.
\newblock An examination of data confidentiality and disclosure issues related
  to publication of empirical ROC curves.
\newblock \emph{Academic Radiology}, 20(7): 889--896.

\bibitem[{McMahan et~al.(2017)McMahan, Moore, Ramage, Hampson, and
  y~Arcas}]{mmr+17}
McMahan, B.; Moore, E.; Ramage, D.; Hampson, S.; and y~Arcas, B.~A. 2017.
\newblock Communication-efficient learning of deep networks from decentralized
  data.
\newblock In \emph{Artificial Intelligence and Statistics}, 1273--1282. PMLR.

\bibitem[{McMahan et~al.(2018)McMahan, Ramage, Talwar, and
  Zhang}]{mcmahan2017learning}
McMahan, H.~B.; Ramage, D.; Talwar, K.; and Zhang, L. 2018.
\newblock Learning Differentially Private Recurrent Language Models.
\newblock In \emph{International Conference on Learning Representations}.

\bibitem[{McSherry and Talwar(2007)}]{dpcomposition2007}
McSherry, F.; and Talwar, K. 2007.
\newblock Mechanism Design via Differential Privacy.
\newblock In \emph{48th Annual IEEE Symposium on Foundations of Computer
  Science (FOCS'07)}, 94--103.

\bibitem[{Menon et~al.(2015)Menon, Rooyen, Ong, and
  Williamson}]{ConvertAUC2015}
Menon, A.; Rooyen, B.~V.; Ong, C.~S.; and Williamson, B. 2015.
\newblock Learning from Corrupted Binary Labels via Class-Probability
  Estimation.
\newblock In Bach, F.; and Blei, D., eds., \emph{Proceedings of the 32nd
  International Conference on Machine Learning}, volume~37 of \emph{Proceedings
  of Machine Learning Research}, 125--134. Lille, France: PMLR.

\bibitem[{Stoddard, Chen, and Machanavajjhala(2014)}]{Dp4ML2014}
Stoddard, B.; Chen, Y.; and Machanavajjhala, A. 2014.
\newblock Differentially Private Algorithms for Empirical Machine Learning.
\newblock \emph{CoRR}, abs/1411.5428.

\bibitem[{Subramanyan et~al.(2017)Subramanyan, Sinha, Lebedev, Devadas, and
  Seshia}]{subramanyan2017formal}
Subramanyan, P.; Sinha, R.; Lebedev, I.; Devadas, S.; and Seshia, S.~A. 2017.
\newblock A formal foundation for secure remote execution of enclaves.
\newblock In \emph{Proceedings of the 2017 ACM SIGSAC Conference on Computer
  and Communications Security}, 2435--2450.

\bibitem[{Sun et~al.(2021)Sun, Yao, Gao, Xie, and Wang}]{sunDefending2021}
Sun, J.; Yao, Y.; Gao, W.; Xie, J.; and Wang, C. 2021.
\newblock Defending against Reconstruction Attack in Vertical Federated
  Learning.
\newblock \emph{CoRR}, abs/2107.09898.

\bibitem[{Vepakomma et~al.(2019)Vepakomma, Gupta, Dubey, and
  Raskar}]{vepakomma2019reducing}
Vepakomma, P.; Gupta, O.; Dubey, A.; and Raskar, R. 2019.
\newblock Reducing leakage in distributed deep learning for sensitive health
  data.
\newblock \emph{arXiv preprint arXiv:1812.00564}.

\bibitem[{Vepakomma et~al.(2018{\natexlab{a}})Vepakomma, Gupta, Swedish, and
  Raskar}]{vgsr18}
Vepakomma, P.; Gupta, O.; Swedish, T.; and Raskar, R. 2018{\natexlab{a}}.
\newblock Split learning for health: Distributed deep learning without sharing
  raw patient data.
\newblock \emph{arXiv preprint arXiv:1812.00564}.

\bibitem[{Vepakomma et~al.(2018{\natexlab{b}})Vepakomma, Gupta, Swedish, and
  Raskar}]{vepakomma2018split}
Vepakomma, P.; Gupta, O.; Swedish, T.; and Raskar, R. 2018{\natexlab{b}}.
\newblock Split learning for health: Distributed deep learning without sharing
  raw patient data.
\newblock \emph{arXiv preprint arXiv:1812.00564}.

\bibitem[{Warner(1965)}]{RR}
Warner, S.~L. 1965.
\newblock Randomized Response: A Survey Technique for Eliminating Evasive
  Answer Bias.
\newblock \emph{Journal of the American Statistical Association}, 60(309):
  63--69.

\bibitem[{Xiong et~al.(2020)Xiong, Liu, Li, Cai, Niu, and
  Del~Rey}]{localdp2020}
Xiong, X.; Liu, S.; Li, D.; Cai, Z.; Niu, X.; and Del~Rey, A.~M. 2020.
\newblock A Comprehensive Survey on Local Differential Privacy.
\newblock \emph{Sec. and Commun. Netw.}, 2020.

\bibitem[{Yang et~al.(2019{\natexlab{a}})Yang, Liu, Chen, and Tong}]{ylct19}
Yang, Q.; Liu, Y.; Chen, T.; and Tong, Y. 2019{\natexlab{a}}.
\newblock Federated machine learning: Concept and applications.
\newblock \emph{ACM Transactions on Intelligent Systems and Technology (TIST)},
  10(2): 1--19.

\bibitem[{Yang et~al.(2019{\natexlab{b}})Yang, Liu, Chen, and
  Tong}]{yang2019federated}
Yang, Q.; Liu, Y.; Chen, T.; and Tong, Y. 2019{\natexlab{b}}.
\newblock Federated machine learning: Concept and applications.
\newblock In \emph{ACM Transactions on Intelligent Systems and Technology
  (TIST)}, 1--19. ACM New York, NY, USA.

\bibitem[{Yuan and Ma(2020)}]{ym20}
Yuan, H.; and Ma, T. 2020.
\newblock Federated Accelerated Stochastic Gradient Descent.
\newblock In Larochelle, H.; Ranzato, M.; Hadsell, R.; Balcan, M.~F.; and Lin,
  H., eds., \emph{Advances in Neural Information Processing Systems},
  volume~33, 5332--5344. Curran Associates, Inc.

\bibitem[{Zhao, Mopuri, and Bilen(2020)}]{zhao2020idlg}
Zhao, B.; Mopuri, K.~R.; and Bilen, H. 2020.
\newblock iDLG: Improved Deep Leakage from Gradients.
\newblock \emph{arXiv preprint arXiv:2001.02610}.

\bibitem[{Zhu, Liu, and Han(2019)}]{zhu2019dlg}
Zhu, L.; Liu, Z.; and Han, S. 2019.
\newblock Deep leakage from gradients.
\newblock In \emph{Advances in Neural Information Processing Systems},
  14774--14784.

\end{thebibliography}




\end{document}